\documentclass[pdflatex,sn-nature]{sn-jnl}
\usepackage{fix-cm}
\usepackage{graphicx}%
\usepackage{multirow}%
\usepackage{amsmath,amssymb,amsfonts}%
\usepackage{amsthm}%
\usepackage{mathrsfs}%
\usepackage[title]{appendix}%
\usepackage{xcolor}%
\usepackage{textcomp}%
\usepackage{manyfoot}%
\usepackage{booktabs}%
\usepackage{algorithm}%
\usepackage{algorithmicx}%
\usepackage{algpseudocode}%
\usepackage{listings}%
\usepackage{hyperref}
\usepackage{float}

\begin{document}

\title[Article Title]{ BlastOFormer: Attention and Neural Operator Deep Learning Methods for Explosive Blast Prediction}

\author[1]{\fnm{Reid} \sur{Graves}}\email{rgraves@andrew.cmu.edu}

\author[1]{\fnm{Anthony} \sur{Zhou}}\email{ayz2@andrew.cmu.edu}

\author*[1]{\fnm{Amir} \sur{Barati Farimani}}\email{barati@cmu.edu}

\affil[1]{\orgdiv{Department of Mechanical Engineering}, \orgname{Carnegie Mellon University}, \orgaddress{\street{5000 Forbes Avenue}, \city{Pittsburgh}, \postcode{15213}, \state{PA}, \country{USA}}}

\abstract{
Accurate prediction of blast pressure fields is essential for applications in structural safety, defense planning, and hazard mitigation. Traditional methods such as empirical models and computational fluid dynamics (CFD) simulations offer limited trade offs between speed and accuracy; empirical models fail to capture complex interactions in cluttered environments, while CFD simulations are computationally expensive and time consuming. In this work, we introduce BlastOFormer, a novel Transformer based surrogate model for full field maximum pressure prediction from arbitrary obstacle and charge configurations. BlastOFormer leverages a signed distance function (SDF) encoding and a grid to grid attention based architecture inspired by OFormer and Vision Transformer (ViT) frameworks. Trained on a dataset generated using the open source blastFoam CFD solver, our model outperforms convolutional neural networks (CNNs) and Fourier Neural Operators (FNOs) across both log transformed and unscaled domains. Quantitatively, BlastOFormer achieves the highest $R^2$ score (0.9516) and lowest error metrics, while requiring only 6.4 milliseconds for inference, more 600,000 times faster than CFD simulations. Qualitative visualizations and error analyses further confirm BlastOFormer’s superior spatial coherence and generalization capabilities. These results highlight its potential as a real time alternative to conventional CFD approaches for blast pressure estimation in complex environments. 
}

\keywords{deep learning, blast loading, neural operator, transformer, attention }

\maketitle
\vspace{0.5cm} 
\noindent\textbf{Article Highlights}
\begin{itemize}
    \item A novel Transformer based architecture, BlastOFormer, is proposed for predicting full field blast pressure distributions.
    \item The model leverages signed distance functions (SDFs) and grid based tokenization to encode obstacle and charge configurations.
    \item BlastOFormer outperforms CNN and FNO baselines across log transformed and unscaled pressure domains in both accuracy and robustness.
    \item Inference time is reduced to 6.4 milliseconds, offering over 600,000 times speedup compared to traditional CFD simulations using blastFoam.
    \item Visual and statistical analyses demonstrate superior generalization and spatial coherence, enabling real time surrogate modeling for blast scenarios.
\end{itemize}

\newpage
\section{Introduction}\label{sec1}

Explosive wave propagation is a highly nonlinear and dynamic phenomenon characterized by complex shock interactions, rapid pressure fluctuations, and significant sensitivity to environmental boundaries and structural geometries. Accurate prediction of blast waves is essential in both military and civilian contexts, informing the design of protective structures, safety protocols, and threat assessment systems. Traditional methods for modeling blast effects typically fall into two categories: empirical and semi-empirical approaches, or high fidelity computational fluid dynamics (CFD) simulations \cite{Filice_anselmo2022}. While empirical models offer rapid approximations, they often oversimplify environmental complexity and neglect structural influences. Conversely, CFD simulations provide detailed and physically accurate solutions but are computationally expensive, time consuming, and must be rerun entirely for each new configuration or scenario \cite{kang_predict, Dennis_1, Li_bleve_machine_learning}.

Empirical models \cite{Filice_anselmo2022, REMENNIKOV20052197, kingery1984air, empirical_review, Beshara1994Modelling}, such as those used to predict peak overpressures from free field explosions, often assume idealized conditions. However, real world settings frequently involve obstacles or confined geometries that drastically alter wave behavior through reflections, diffractions, and coalescence. Remennikov et al. \cite{REMENNIKOV20052197} demonstrated using CFD that surrounding structures can increase reflected pressures by up to 300\% compared to open field conditions, emphasizing the need for more adaptable and precise modeling strategies in built environments.

Computational methods \cite{Filice_anselmo2022, Abdoh2024, Talaat2022, Dib2022} address the shortfalls of empirical models by leveraging physical models in simulation to account for complex geometries and wave reflection effects. However, these computational approaches are computationally expensive due to the high speeds and large deformations inherent to blast wave scenarios, requiring dynamic mesh adaptation \cite{Abdoh2024}. Furthermore, such simulations are predominantly implemented using proprietary software \cite{Abdoh2024}.

To address the computational burden of CFD, several recent studies have explored the use of machine learning to develop fast surrogate models. Dennis et al. \cite{Dennis_1} trained a two layer neural network on 72 blast simulations with varying charge sizes and locations, predicting peak specific impulse at discrete points with an error rate of approximately 10\% relative to CFD. Building on this, Dennis et al. \cite{dennis2} proposed a dual network strategy to separately model obstructed and unobstructed zones, reducing mean absolute error to 5 kPa and achieving 93\% accuracy in predicting eardrum rupture outcomes, while offering a 30 times speedup over CFD.

Li et al. \cite{LI_Comparative_study} expanded the scope of model evaluation by benchmarking traditional and modern machine learning architectures including decision trees, multi layer perceptrons (MLPs), ResNets \cite{he2015deepresiduallearningimage}, and Transformers \cite{vaswani2023attentionneed} on blast scenarios involving boiling liquid expanding vapor explosions (BLEVEs). Their results highlighted the Transformer model’s superior accuracy, with a mean absolute percentage error of just 3.5\%, underscoring the potential of attention based architectures for modeling complex physical systems. Furthermore, there have been many architectures and methods proposed for PDE deep learning \cite{li2023scalable, shu2023physics, li2023transformer, lorsung2023mesh, zhou2024predicting}.

Despite these advancements, existing approaches predominantly focus on point-wise pressure predictions using relatively simple or shallow networks, which constrains their ability to generalize across spatial domains and capture complex phenomena such as shock reflections and obstacle-induced interactions. To address these limitations, we propose a model that simultaneously predicts pressure fields across the entire spatial domain, leveraging advanced architectures that effectively learn intricate spatial dependencies and environmental conditions. This architectural approach enables significantly faster pressure prediction compared to point-wise techniques that require numerous forward passes for a single environment. Furthermore, our model demonstrates capability in predicting pressure mappings with arbitrary obstacle configurations, enhancing its applicability to diverse real world scenarios.

Specifically, we introduce BlastOFormer, a novel Transformer based architecture adapted from the Operator Transformer (OFormer) framework \cite{li2023transformer}, with input patching strategies inspired by the Vision Transformer (ViT) \cite{dosovitskiy2020vit}. We compare BlastOFormer to convolutional neural networks (CNNs) and Fourier Neural Operators (FNOs) \cite{neural_operator}, evaluating each on a large scale dataset generated using the open source blastFoam CFD solver. Our findings demonstrate that BlastOFormer achieves superior accuracy and generalization, while enabling rapid, full field pressure predictions suitable for real time applications.

\section{Methodology}
\begin{figure}[h]
    \centering
    \includegraphics[width=\textwidth]{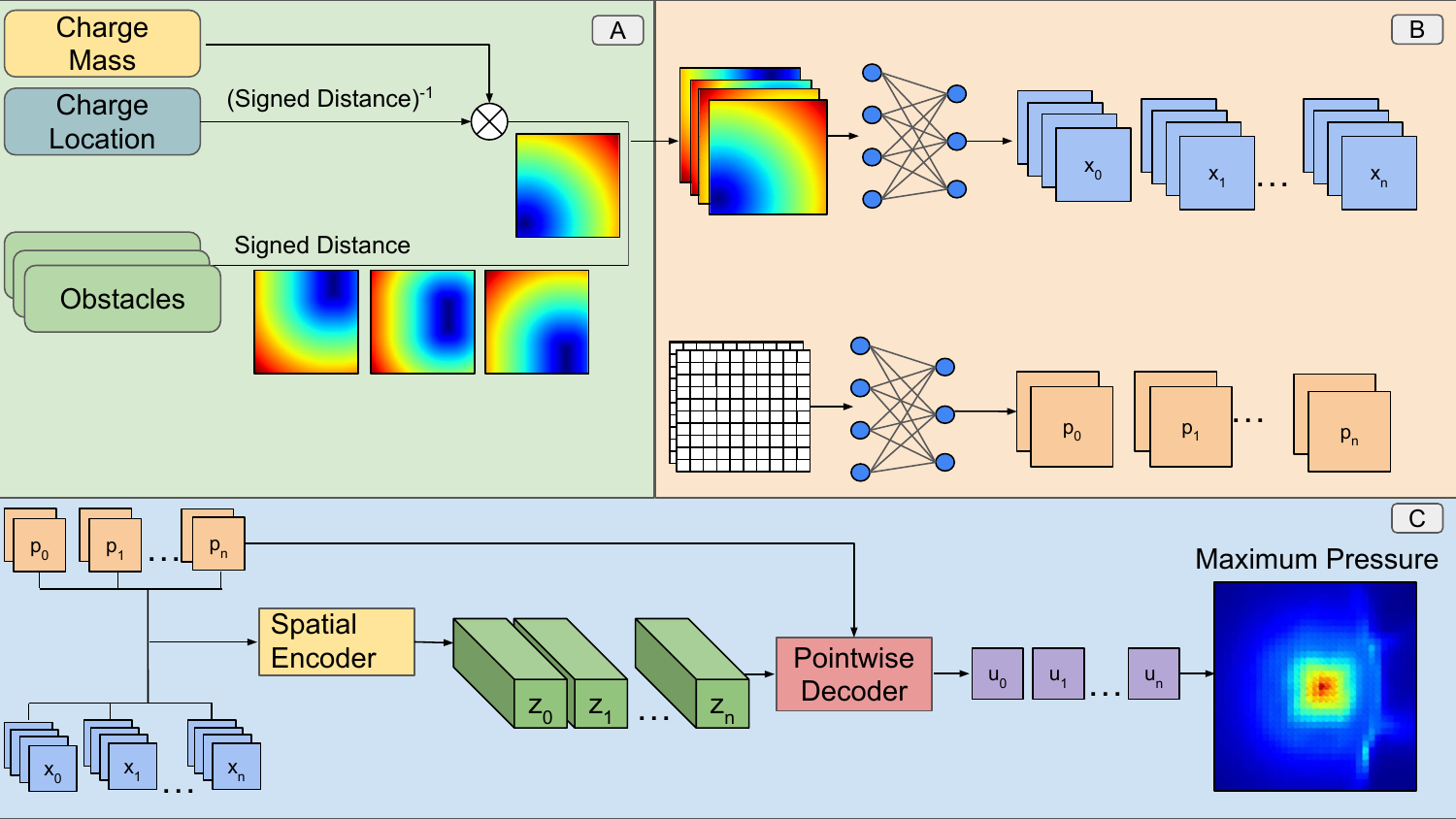}
    \caption{High level overview of the BlastOFormer architecture. 
    (A) The input data, consisting of charge mass, charge locations, and obstacle positions, is transformed into a grid based, multi channel format. Charge mass is encoded into a channel by multiplying it with the inverse signed distance function, while each obstacle is encoded into a separate channel representing its signed distance.
    (B) Inspired by Vision Transformer (ViT), a linear patching layer projects this multi channel grid input into sequences of value and positional tokens. Spatial coordinates for both input and query points undergo a similar patch wise linear transformation.
    (C) The spatial encoder generates high dimensional embeddings capturing spatial features and input relationships. These embeddings ($z_0, z_1, \dots, z_n$), combined with the encoded coordinate tokens, feed into a point wise decoder. The decoder generates prediction tokens, which are subsequently reassembled into the predicted maximum pressure field through a learned linear depatchification layer.}
    \label{blastOFormer_model}
\end{figure}

\subsection{Dataset Creation}

\begin{figure}
    \centering
    \includegraphics[width=\linewidth]{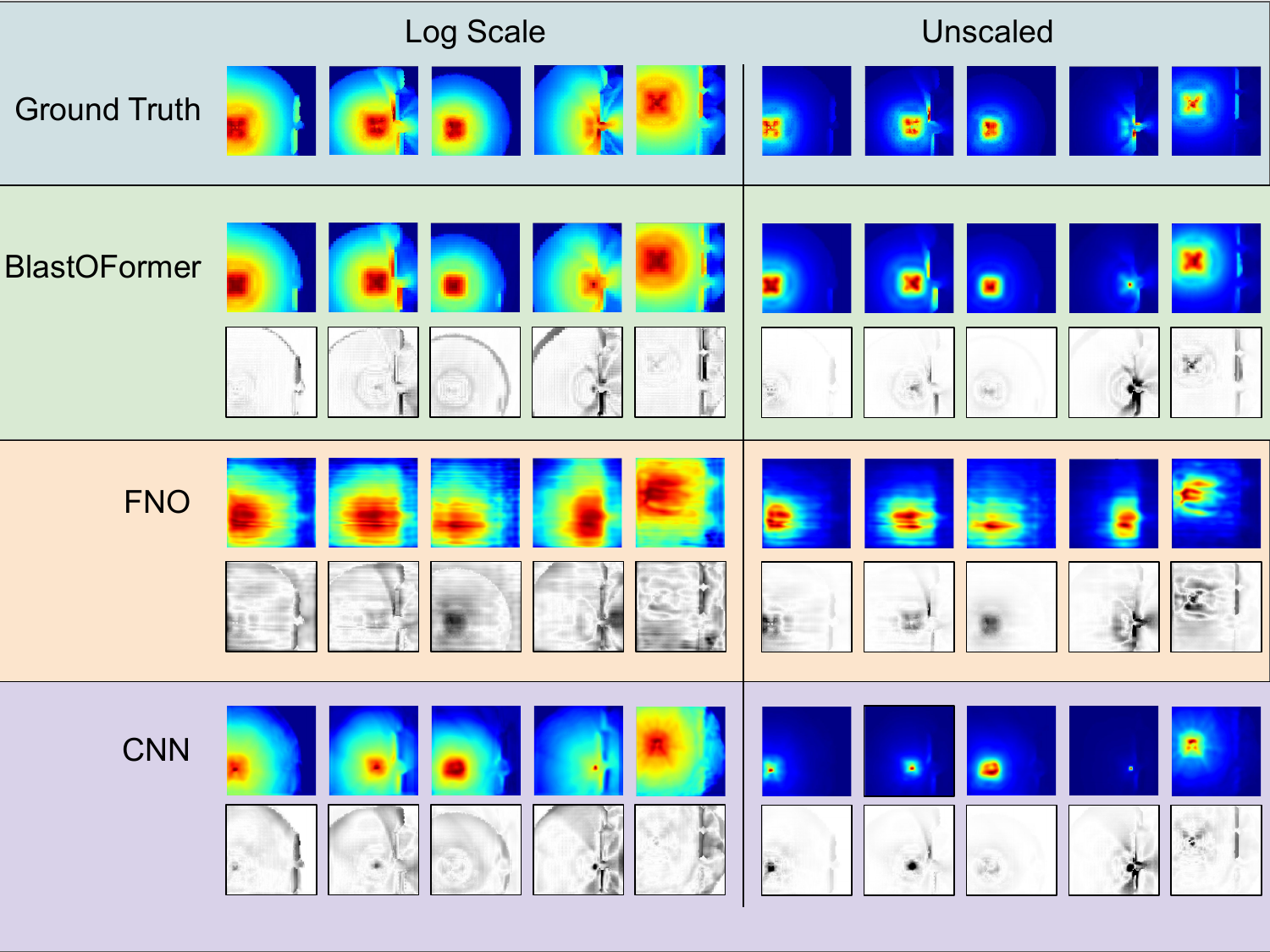}
    \caption{The top row displays ground truth pressure maps generated using BlastFoam. The first, second, and third rows show pressure predictions from BlastOFormer, FNO, and a CNN model, respectively. Each column represents a different pressure domain: the left column shows log-scaled pressures, while the right column shows unscaled values. For each model, the corresponding error maps—shown beneath the predictions with greyscale coloring highlight the differences from the ground truth. All three models successfully capture the key features of the BlastFoam simulations. The visualizations use the Jet colormap, which transitions from blue (low values) to red (high values). In particular, the log transformed pressure distributions demonstrate smooth gradients radiating outward from the charge center, consistent with expected behavior for log transformed data.}
    \label{fig:enter-label}
\end{figure}


For model training and validation, we developed a comprehensive dataset comprising 1500 samples generated using blastFoam, a specialized detonation simulation library built upon the open source computational fluid dynamics (CFD) platform OpenFOAM~\cite{blastfoam,heylmun_blastfoamguide_2021,openfoam}. Our simulation environment adapts a scenario from a publicly available blastFoam tutorial~\cite{blastfoam}, configured with the critical parameters detailed in Fig.~\ref{blastfoam} in the appendix and summarized below. We employed a systematic data partitioning strategy with 1100 training samples (73.33\%), 200 validation samples (13.33\%), and 200 test samples (13.33\%) to ensure robust model evaluation while maintaining sufficient training data for complex pattern recognition.

\begin{itemize}
\item \textbf{Simulation Domain:}
The simulation environment is a cubic domain measuring $10 m \times 10 m \times 10 m$, partitioned into 25 blocks along both the x and y axes and 10 blocks along the z axis. The domain is symmetrically centered at the origin, spanning coordinates from $-5m$ to $5m$ in all three spatial dimensions.
\item \textbf{Pressure Probes:}
Pressure measurements are recorded using a uniform grid of probes positioned at a constant height of $1\,m$ along the z axis. Probes are evenly spaced every $0.1m$ across the x and y axes, covering coordinates from $-4.9m$ to $4.9m$, thus forming a $99 \times 99$ grid. This arrangement yields $9,801$ individual pressure readings at each timestep of the simulation.

\item \textbf{Charge Configuration:}
In each simulation run, the explosive charge is randomly placed within coordinates $-4.9 \leq x \leq 4.9$ and $-4.9 \leq y \leq 2.0$. The charge mass ($c_m$) is also randomly selected, ranging between $5kg$ and $50kg$.

\item \textbf{Obstacle Configuration:}
Each simulation scenario includes three obstacles, with randomized positions defined as follows:
\begin{enumerate}
    \item \textbf{Obstacle 1:} $-4.9m \leq x_{min} \leq -4.5m$, and $-2.5m \leq x_{max} \leq 2.25m$
    \item \textbf{Obstacle 2:} $-2.0m \leq x_{min} \leq -1.9m$, and $1.0m \leq x_{max} \leq 1.5m$
    \item \textbf{Obstacle 3:} $1.5m \leq x_{min} \leq 2.5m$, and $4.5m \leq x_{max} \leq 4.9m$
\end{enumerate}
All three obstacles share a common y range, randomly set between $2.0,m \leq y_{min} \leq 3.0m$, with their width determined by $y_{max} = y_{min} + h_y$, where $0.5m \leq h_y \leq 1.0m$. The obstacles extend vertically from $z_{min} = 0m$ to $z_{max} = z_{min} + h_z$, with the height $h_z$ randomly chosen from $0.5m$ to $2.0m$. 
\item \textbf{Numerical Configuration:}
Our simulations employed the blastFoam solver configured with second-order accurate discretization schemes, specifically least squares gradient schemes, Riemann divergence schemes, and vanAlbada reconstruction schemes for improved accuracy. Time integration was performed using an Euler scheme. The simulation was executed in parallel across 8 processors using OpenFOAM's decomposition utility. We employed a timestep size of $1\times10^{-7},s$ with a total simulation duration of $7.5\times10^{-6}s$, yielding approximately $450$ total timesteps per simulation, and solver tolerances set to $1\times 10^{-6}$. Mesh refinement was conducted using the snappyHexMesh utility to adaptively capture relevant shock phenomena. Each simulation typically required approximately 10-12 minutes on an Intel Core i7-14700Fx28 processor.

\end{itemize}

\subsection{BlastOFormer}

\begin{figure}[h]
    \centering
    \includegraphics[width=\linewidth]{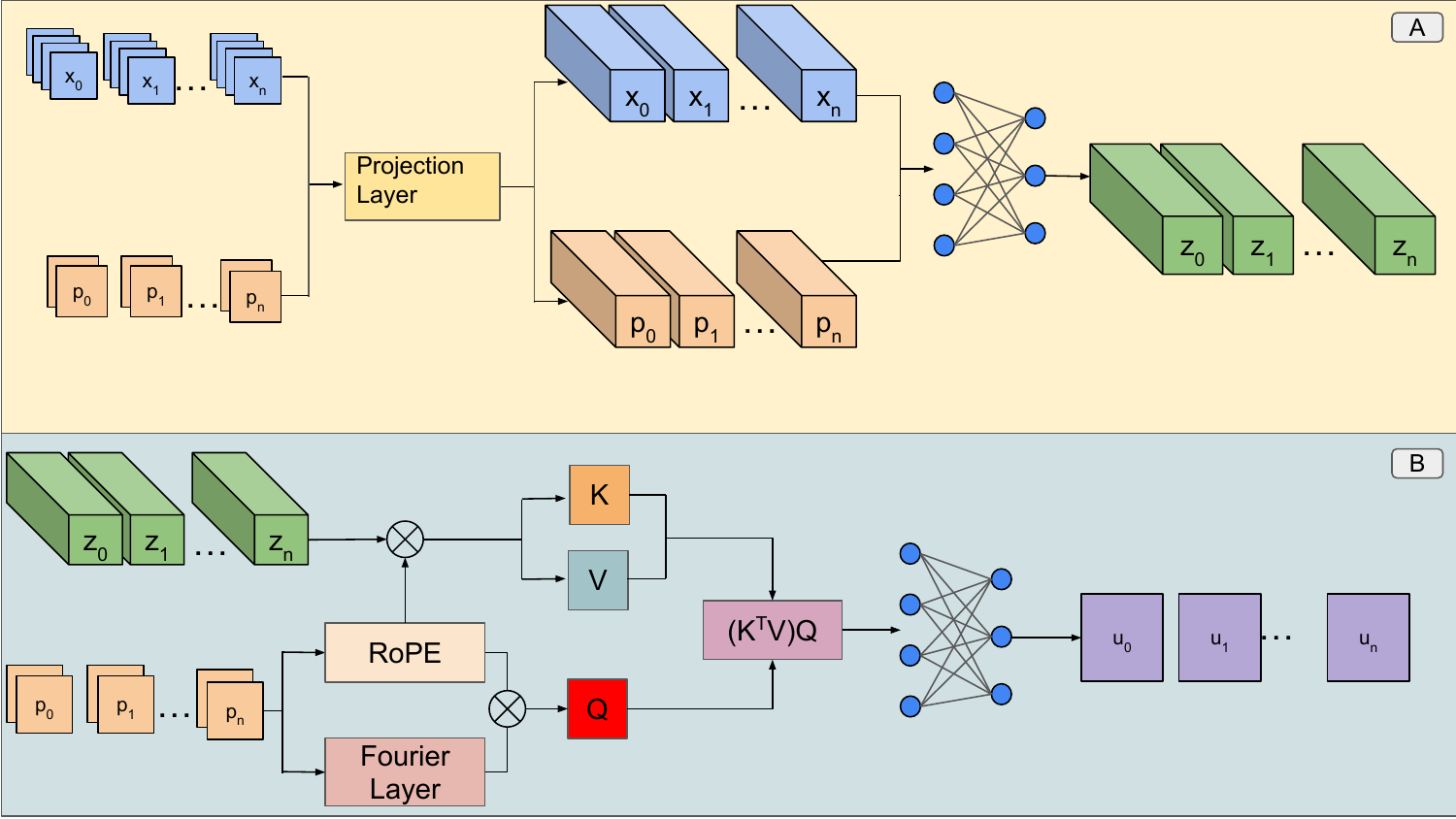}
    \caption{The encoder and decoder components from figure \ref{blastOFormer_model}.  (A)~The encoder processes signed distance values ($x_0,x_1,\dots, x_n$) and positional information ($p_0, p_1, \dots, p_n$) pass through a projection layer, expanding the environmental features before being combined into latent initial condition tokens ($z_0, z_1, \dots, z_n$). The decoder is detailed in (B), pictorially visualizing the operations involved to combine initial conditional tokens and output pressure mapping coordinates ($p_0, p_1, \dots, p_n$) to formulate the query, key and value vectors for the attention operation, followed by a linear layer to transform the predicted pressure values back to physical dimensions, outputing the predicted pressure values $(u_0, u_1, \dots, u_n)$.}
    \label{encoder_decoder}
\end{figure}

Our BlastOFormer architecture draws inspiration from the Operator Transformer (OFormer)~\cite{li2023transformer} and Vision Transformer (ViT)~\cite{dosovitskiy2020vit} frameworks, specifically tailored for predicting maximum pressure fields from charge and obstacle configurations. We selected this transformer based approach based on the hypothesis that its attention mechanisms would more effectively capture complex relationships between environmental parameters (charge mass, obstacle positioning) and dynamic phenomena (wave reflections, interactions) compared to traditional techniques such as Convolutional Neural Networks. The transformer's inherent ability to model long range dependencies is particularly advantageous for blast propagation scenarios where pressure effects can transmit across significant distances and interact with multiple environmental elements. The model pipeline consists of the following key steps:

\begin{enumerate}
\item The charge mass, charge locations, and obstacle positions are processed through a signed distance function to generate a multi channel grid based input tensor.
\item This input tensor, along with its spatial grid coordinates, undergoes linear patch wise projections, resulting in value and positional token sequences suitable for the spatial encoder.
\item The spatial encoder employs linear projections to embed these tokens into a high dimensional feature space.
\item A stack of self attention blocks, augmented by Rotary Positional Embeddings (RoPE) \cite{RoFormer}, captures both the local function values and spatial relationships, producing enriched input embeddings.
\item The decoder takes the encoder's high dimensional features, together with positional tokens, to reconstruct the target spatial representation.
\item Query location tokens undergo Fourier Feature projection transformations to enrich spatial positional encoding.
\item Cross attention layers integrate the query tokens with the encoder embeddings, effectively capturing complex interactions at each query location.
\item Finally, a linear depatchification layer converts these tokens back into a structured, two dimensional, single channel output tensor, representing the predicted maximum pressure field.
\end{enumerate}

Each component of BlastOFormer is detailed further in the subsequent sections.

\subsubsection{Model Input Formatting}

Architectures such as Oformer \cite{li2023transformer} and ViT \cite{dosovitskiy2020vit} employ grid to grid input to output mappings. Inspired by this design, our proposed BlastOFormer architecture similarly adopts a structured input format, where each query output point uniquely corresponds to specific input features. This ensures clear associations between input conditions and predicted outcomes, allowing the model to effectively learn precise mappings from environmental conditions and configurations.

To encode obstacle and charge characteristics into a consistent representation, we leverage a Signed Distance Function (SDF) \cite{Osher2003, level_set}, defined as:

\begin{align*}
f(x) = \begin{cases}
d(x, \partial \Omega) & \text{if } x \in \Omega \\
-1 & \text{if } x \notin \Omega
\end{cases}
\end{align*}

Here, $\Omega$ denotes the spatial region external to obstacles or charges, $x$ is the spatial position evaluated, and $d(x, \partial \Omega)$ represents the minimum distance from point $x$ to the boundary $\partial \Omega$, indicating obstacle or charge locations.

As illustrated in Fig. \ref{blastOFormer_model} (A), each obstacle is represented by a separate tensor channel. Charge properties, including mass and spatial location, are encoded in a combined single channel. Specifically, the charge mass is modulated by the inverse of the signed distance, effectively capturing increased pressure near the charge center and decreasing pressure further away.

Consistent with the original formulation of OFormer \cite{li2023transformer}, our approach distinguishes between the function values $a(x_i)$ and their associated input coordinates $x_i$. This distinction is visually represented in Fig. \ref{blastOFormer_model} (B). Given that our problem is inherently two dimensional, the input coordinates are structured as a two channel tensor with spatial dimensions of $99 \times 99$.

\subsubsection{Image Patchification}
The input to our model, consisting of function values $a(x_i)$ and their corresponding coordinates $x_i$, can be conceptually compared to RGB image data, where an image is divided into three distinct channels representing red, green, and blue intensities. Similarly, our function values $a(x_i)$ are segmented into four channels: three channels represent the signed distance for each obstacle, while the fourth channel encodes the inverse signed distance of the charge scaled by its mass. Additionally, our input coordinates are represented by two separate channels corresponding to the spatial $x$ and $y$ locations.

Given the considerable spatial dimensions ($99 \times 99$) of our input tensors, direct processing would be computationally intensive. Inspired by the approach introduced in \cite{dosovitskiy2020vit}, we address this challenge by partitioning our input data into smaller patches. Specifically, we employ learned linear projection layers that transform the function value tensor (initially shaped $[\text{batch size}, 99, 99, 4]$) and the coordinate tensor (initially shaped $[\text{batch size}, 99, 99, 2]$) into token sequences of shapes $[\text{batch size}, \frac{9801}{(\text{patch size})^2}, 4]$ and $[\text{batch size}, \frac{9801}{(\text{patch size})^2}, 2]$, respectively. This strategy effectively reduces computational demands while preserving crucial spatial and feature information.

\subsubsection{Spatial Encoder}

The spatial encoder architecture is detailed in Fig. \ref{encoder_decoder} (A). Initially, the input function values $a(x_i)$ and corresponding spatial locations $x_i$ are partitioned into patches and tokenized. These tokens are then projected into a higher dimensional feature space using learned linear layers. The function value tokens undergo layer normalization \cite{ba2016layernormalization} before being processed through multiple self attention blocks, which operate jointly on both the function values and spatial coordinates. Following the self attention, another layer normalization is performed. The final output from the spatial encoder is a set of enriched feature vector tokens $z_i$, which effectively represent the complex relationships between the input function values and spatial coordinates.

\subsubsection{Pointwise Decoder}

The pointwise decoder architecture is illustrated in Fig. \ref{encoder_decoder} (B). Central to the decoder is a cross attention mechanism \cite{gheini2021crossattentionneedadaptingpretrained, cross_attention_network, crossViT} designed to effectively capture relationships among input function values, input spatial coordinates, and query locations. Although our architecture currently shares query location coordinates with input spatial coordinates, this constraint is not strictly required.

The decoder leverages Random Fourier Feature (RFF) projection \cite{fourier_feature_projection, tran2023factorizedfourierneuraloperators, tancik2020fourierfeaturesletnetworks} applied to query coordinates, defined mathematically as:
\begin{align*}
    \gamma(Y) = [\cos(2\pi Y B), \sin(2\pi Y B)]
\end{align*}

where $Y = [y_1, y_2, \dots, y_n]^T$ are the coordinate tokens and $B \in \mathbb{R}^{d_1 \times d_2}$ is sampled from a Gaussian distribution $\mathcal{N}(0,\sigma^2)$, as detailed in \cite{li2023transformer}. This transformation encodes coordinates into high-dimensional sinusoidal features, enabling the model to represent complex, high frequency variations of the target function.

In the cross attention step, the keys ($K$) and values ($V$) are derived from the feature vector tokens $z_i$, while the query ($Q$) originates from the query locations. Given the permutation invariance of attention mechanisms \cite{lee2019settransformerframeworkattentionbased}, Rotary Positional Embedding (RoPE) \cite{RoFormer} is employed. RoPE applies a rotation via a block diagonal matrix $\Theta(x_i, x_j)$ to $K, V, Q$, embedding the relative spatial offsets between points $i$ and $j$ into the attention calculation. Consequently, RoPE enhanced keys, values, and queries ($K', V', Q'$) carry precise spatial information.

The enriched features obtained through cross-attention are computed as:
\begin{align*}
    z' = z + \text{CrossAttn}(Q',K',V')
\end{align*}
and subsequently processed by a feed-forward linear layer to yield the final output tokens $u_i$, effectively translating the learned spatial contextual representations into meaningful predictions.
Finally, the output tokens $u_i$ are passed through a learned linear depatchification layer, outputting the predicted maximum pressure grid.

\subsubsection{Scaling Convolutional Neural Network}

To improve fidelity in the unscaled pressure domain, we introduce an auxiliary convolutional neural network that learns a residual mapping from log transformed model predictions to their corresponding unscaled values. The network is trained post hoc using log domain predictions from the trained BlastOFormer model. Specifically, predicted log pressure maps are passed through the UnscalerCNN, which is trained to minimize the mean squared error with respect to the unscaled blastFoam ground truth data. The architecture consists of six convolutional layers with progressively increasing and then decreasing channel dimensions, forming an encoder-decoder structure. All layers use $3 \times 3$ kernels with padding to preserve spatial dimensions, and ReLU activations are applied between layers. This lightweight correction network enhances the model's quantitative accuracy in the physical unscaled domain without modifying the original pressure predictor.

\subsection{Model Configuration and Training Parameters}
\subsubsection{Model Hyperparameters}
The BlastOFormer architecture was configured with the following hyperparameters:
\begin{itemize}
    \item \textbf{Encoder Configuration}:
    The spatial encoder employs an input embedding dimension of 96 channels, which are then projected to a sequence embedding dimension of 256. The encoder consists of 6 transformer layers with 4 attention heads each, operating on a spatial resolution of $99\times99$.
    \item \textbf{Decoder Configuration}
    The decoder maintains latent channels of 256 dimensions to match the encoder output, ultimately producing single channel pressure predictions at the same $99\times99$ resolution.
    \item \textbf{Patch Size}: We examined patch sizes 11, 9, 3 and 1, to ensure divisibility of the pressure grid by the patches. We ultimately selected a patch size of 1 during model comparison to baselines.
\end{itemize}

\subsubsection{Training Configuration}
The model was trained using the following optimization strategy:
\begin{itemize}
    \item \textbf{Optimizer}: AdamW optimizer with a learning rate of $1\times10^{-4}$ and default $\beta$ parameters.
    \item \textbf{Learning Rate Scheduler}: Cosine annealing schedule over the full training duration.
    \item \textbf{Loss Function}: L1 loss (Mean Absolute Error) was used as the primary training objective. This provide more stable training, given the large magnitude of errors from pressure predictions, preventing squaring of large error terms.
    \item \textbf{Batch Size}: 4 samples per batch when using patch size of 1, due to GPU memory restrictions.
    \item \textbf{Training Duration}: Maximum of 10,000 epochs with early stopping based on validation loss, with early stopping after 1000 epochs with no improvement.
\end{itemize}

The model was trained on a single NVIDIA GPU, with training progress monitored using both TensorBoard and Weights \& Biases (wandb) for experiment tracking. The best model checkpoint was selected based on validation loss performance.

\subsection{Baseline Models}

To evaluate the effectiveness of BlastOFormer, we compare it against two established deep learning architectures that represent different approaches to spatial prediction tasks.

\subsubsection{Convolutional Neural Network (CNN)}

We selected a CNN as our first baseline due to its prevalence in existing blast wave analysis literature and its proven effectiveness in spatial prediction tasks. CNNs have been the de facto standard for image based regression problems and represent the most straightforward deep learning approach for our domain.

The CNN architecture was configured with the following specifications:
\begin{itemize}
    \item \textbf{Architecture}: Six convolutional layers with ReLU activations
    \item \textbf{Base dimension}: 128 channels
    \item \textbf{Model size}: 2.96 million parameters
    \item \textbf{Training configuration}:
    \begin{itemize}
        \item Optimizer: AdamW with learning rate $1 \times 10^{-3}$
        \item Batch size: 32
        \item Learning rate schedule: Cosine annealing
        \item Loss function: L1 loss
        \item Early stopping patience: 100 epochs
    \end{itemize}
\end{itemize}

The CNN processes the signed distance function inputs directly through convolutional operations, learning local spatial patterns through its hierarchical feature extraction. This architecture serves as a strong baseline representing traditional deep learning approaches to spatial regression.

\subsubsection{Fourier Neural Operator (FNO)}

As our second baseline, we implemented a Fourier Neural Operator (FNO) to provide a comparison with another operator based framework. FNOs have demonstrated state of the art performance on various PDE related tasks and represent a fundamentally different approach from CNNs by learning mappings in the frequency domain.

The FNO configuration includes:
\begin{itemize}
    \item \textbf{Architecture parameters}:
    \begin{itemize}
        \item Fourier modes: 6 $\times$ 6 (modes1 $\times$ modes2)
        \item Width: 24 channels
        \item Number of layers: 4
        \item Conditioning channels: 21 (for obstacle and charge encoding)
    \end{itemize}
    \item \textbf{Model size}: 365,000 parameters
    \item \textbf{Training configuration}:
    \begin{itemize}
        \item Optimizer: AdamW with learning rate $1 \times 10^{-4}$
        \item Batch size: 32
        \item Learning rate schedule: Cosine annealing
        \item Loss function: L1 loss on normalized pressure values
        \item Early stopping patience: 100 epochs
    \end{itemize}
\end{itemize}

The FNO processes the charge mass as the primary input while incorporating obstacle positions through a conditioning mechanism. This architecture learns global basis functions in the Fourier domain, enabling it to capture long range dependencies more effectively than purely local convolutional operations.

\subsubsection{Rationale for Baseline Selection}

These baselines were chosen to represent two fundamentally different approaches to spatial function approximation:

\begin{enumerate}
    \item \textbf{CNN}: Represents the traditional deep learning approach with local receptive fields and hierarchical feature learning. This baseline establishes the performance of standard computer vision techniques applied to our physics-based regression task.
    
    \item \textbf{FNO}: Represents modern operator learning methods that work in the frequency domain. As an operator based approach like BlastOFormer, it provides a direct comparison of different strategies for learning function to function mappings.
\end{enumerate}

Together, these baselines span the spectrum from local (CNN) to global (FNO) processing strategies, allowing us to evaluate whether BlastOFormer's attention based mechanism offers advantages over both approaches. We note that the substantially smaller size of the FNO model compared to BlastOFormer and the CNN models was intentional. We explored larger FNO models, but found that predictive performance decreased when increasing it's size.

\section{Results and Discussion}
In this section we present the benchmark results on the dataset generated using blastFoam \cite{blastfoam}, and compare it with baseline architectures using FNO \cite{neural_operator} and CNN architectures.

\subsection{Quantitative Performance}
\begin{table}[h]
    \centering
    \begin{tabular}{cccccc}
    \hline
       Model  & $R^2$  & MAE  & MAPE (\%) &  prediction time (ms) & \# Parameters\\
       \hline
        \textbf{BlastOFormer} & & & & 6.4 & $2.43 \times 10^6$ \\
        log & 0.9169 & \textbf{0.1729} & \textbf{1.315} & &\\
        unscaled & \textbf{0.9516} & \textbf{484 kPa} & \textbf{21.1}& &\\
        \hline
        \textbf{FNO} & & & & 4.0 & $3.65 \times 10^5$\\
        log & 0.8231 & 0.3359 & 2.579 & &\\
        unscaled & 0.9164 & 600 kPa & 35.9& &\\
        \hline
        \textbf{CNN} & & & & \textbf{1.4} & $2.96\times10^6$\\
        log & \textbf{0.9218} & 1.116 & 8.53 & &\\
        unscaled & 0.8945 & 1192 kPa & 211 & &\\
        \hline
    \end{tabular}
    \caption{Performance metrics BlastOFormer, FNO and CNN models as compared to the blastFoam generated samples.}
    \label{performance_table}
\end{table}

Table \ref{performance_table} summarizes the predictive performance of BlastOFormer, FNO, and CNN models, evaluated using the coefficient of determination ($R^2$), mean absolute error (MAE), and average percentage error, both in log transformed and unscaled pressure spaces.

Across all metrics, BlastOFormer consistently outperforms the other models, particularly in the unscaled domain. It achieves the highest $R^2$ score of 0.9516, the lowest unscaled MAE of 484 kPa, and the smallest average percentage error of 21.1\%. These results demonstrate BlastOFormer's superior ability to accurately reconstruct the pressure distribution magnitudes, not just the qualitative features.

In the log-transformed domain, which emphasizes spatial distribution rather than absolute magnitude, the CNN achieves a slightly higher $R^2$ score (0.9218) than BlastOFormer (0.9169). However, its MAE and percentage error are significantly worse, indicating it is less reliable overall. This discrepancy suggests that while the CNN may approximate the general shape of the pressure field, it struggles with accurate scaling and amplitude prediction, which is particularly evident in its unscaled performance, where the average error exceeds 200\%.

FNO performs better than CNN in the unscaled domain but is still outperformed by BlastOFormer in all metrics. Notably, FNO's percentage error in the unscaled domain is 35.9\%, which is nearly 70\% higher than that of BlastOFormer, suggesting a less accurate modeling of the nonlinear effects governing the blast pressure field.

Overall, these results highlight the effectiveness of the joint encoding of spatial coordinates and physical features used in BlastOFormer, allowing it to generalize well across a wide range of blast scenarios. The clear gap between log transformed and unscaled performance further emphasizes the importance of evaluating models in both domains to avoid misleading conclusions based solely on normalized or transformed representations. In addition to its superior predictive accuracy, BlastOFormer also offers a practical advantage in computational efficiency. As shown in Table \ref{performance_table}, BlastOFormer generates full field pressure predictions in just 6.4 milliseconds, which is orders of magnitude faster than the blastFoam simulations, which require approximately 10–12 minutes per sample on an 8 core CPU. This represents a speedup exceeding 5 orders of magnitude, enabling near instantaneous pressure estimation once the model is trained. While the CNN achieves the fastest inference time at 1.4 milliseconds, it does so with significantly higher error, particularly in the unscaled domain. BlastOFormer thus presents the best trade off between speed and accuracy, making it well suited for real time or large scale simulation scenarios where traditional CFD methods are computationally prohibitive.

\subsection{Pressure Map Analysis}
\begin{figure}[h]
    \centering
    \includegraphics[width=\linewidth]{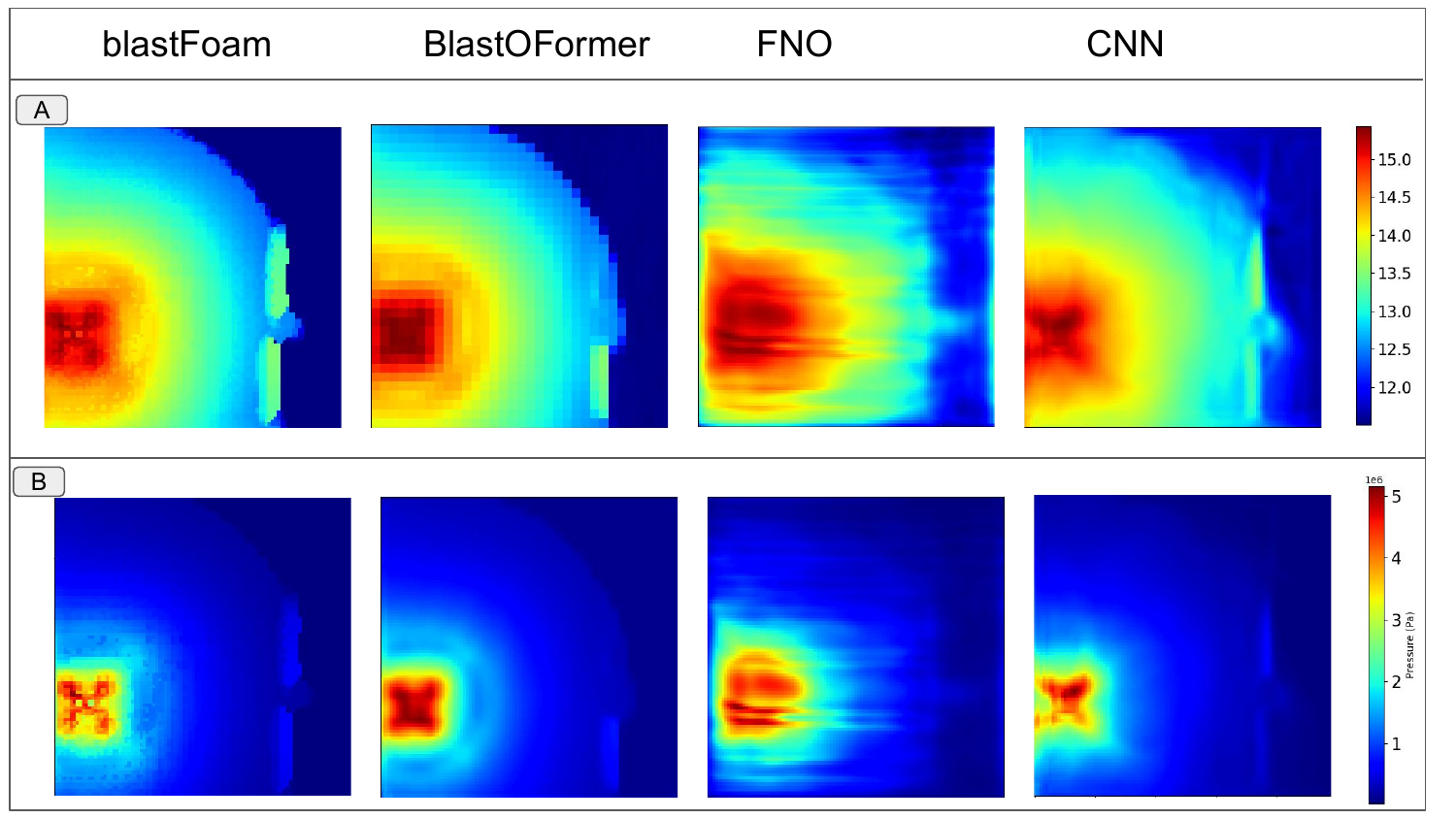}
    \caption{(A) From left to right, the log transformed maximum pressure maps generated by blastFoam (ground truth), BlastOFormer, FNO, and CNN. (B) Corresponding unscaled maximum pressure maps from blastFoam, BlastOFormer, FNO, and CNN. All three models effectively capture the essential features present in the ground truth provided by blastFoam. The visualizations utilize the Jet color map, transitioning from blue (lower values) to red (higher values). The log transformed pressure distributions exhibit a smoother gradient from red to blue as the distance from the charge center increases, aligning well with expected physical behavior.}
    \label{pressure_maps}
\end{figure}

Figure \ref{pressure_maps} provides a visual comparison between the predicted maximum pressure fields from the three models and the ground truth simulation data from blastFoam. Subfigure (A) displays the log transformed pressure maps, while subfigure (B) shows the corresponding unscaled maps. These visualizations provide insight into how well each model captures both the spatial structure and intensity of blast pressures.

In the log transformed domain, all models successfully capture the qualitative behavior of the pressure field, notably the radial decay pattern centered around the charge. Among the three, BlastOFormer most closely replicates the ground truth, maintaining a smooth and continuous transition from the charge center outward. Both FNO and CNN show reasonable agreement with the ground truth, but their outputs exhibit less consistent radial smoothness. FNO in particular struggles to preserve the gradual decay of pressure, instead producing elevated values near the charge center and less coherent gradients across space. While the CNN model retains smoother transitions than FNO, it gradually loses the radial symmetry of the pressure distribution at greater distances from the charge.

BlastOFormer also accurately predicts the increased pressure above the lowermost obstacle, though it underestimates this effect for the central obstacle. Interestingly, CNN appears to better capture this localized pressure increase across both obstacles. In contrast, FNO fails to represent the pressure amplification near either obstacle but does capture some of the increased pressure between obstacles, likely due to wave reflection effects. Overall, while all models capture the general decay trend, BlastOFormer provides the most coherent and physically consistent representation of the log transformed pressure field.

In the unscaled pressure domain, similar trends are observed across the models, with BlastOFormer again demonstrating the closest resemblance to the ground truth. Its predictions maintain the qualitative structure of the blast field, including the radial decay and elevated pressure zones, while slightly overestimating pressure magnitudes both near the charge center and in the far field. This slight overprediction is considered a favorable trait, as it results in more conservative pressure estimates without compromising alignment with the underlying physics. BlastOFormer also successfully captures the increased pressure over the lowermost obstacle, though it continues to underrepresent this effect for the central obstacle.

FNO, consistent with its behavior in the log-transformed domain, struggles to preserve the smooth pressure decay and exhibits reduced spatial coherence, particularly in high gradient regions. CNN similarly fails to retain the symmetry of the pressure field, and this limitation becomes more pronounced in the unscaled domain, especially near the charge center where its predictions deviate significantly from the ground truth. While CNN continues to detect elevated pressures over both the lower and middle obstacles, these are largely confined to the leading edges. In contrast, BlastOFormer more accurately captures the extent of elevated pressure past the leading edge, offering a more realistic depiction of pressure propagation around solid bodies. Overall, the unscaled domain reveals a sharper distinction in model fidelity, reinforcing BlastOFormer's advantage in reproducing both the magnitude and structure of the pressure field.

\subsection{Error Map Analysis}

\begin{figure}[h]
    \centering
    \includegraphics[width=\linewidth]{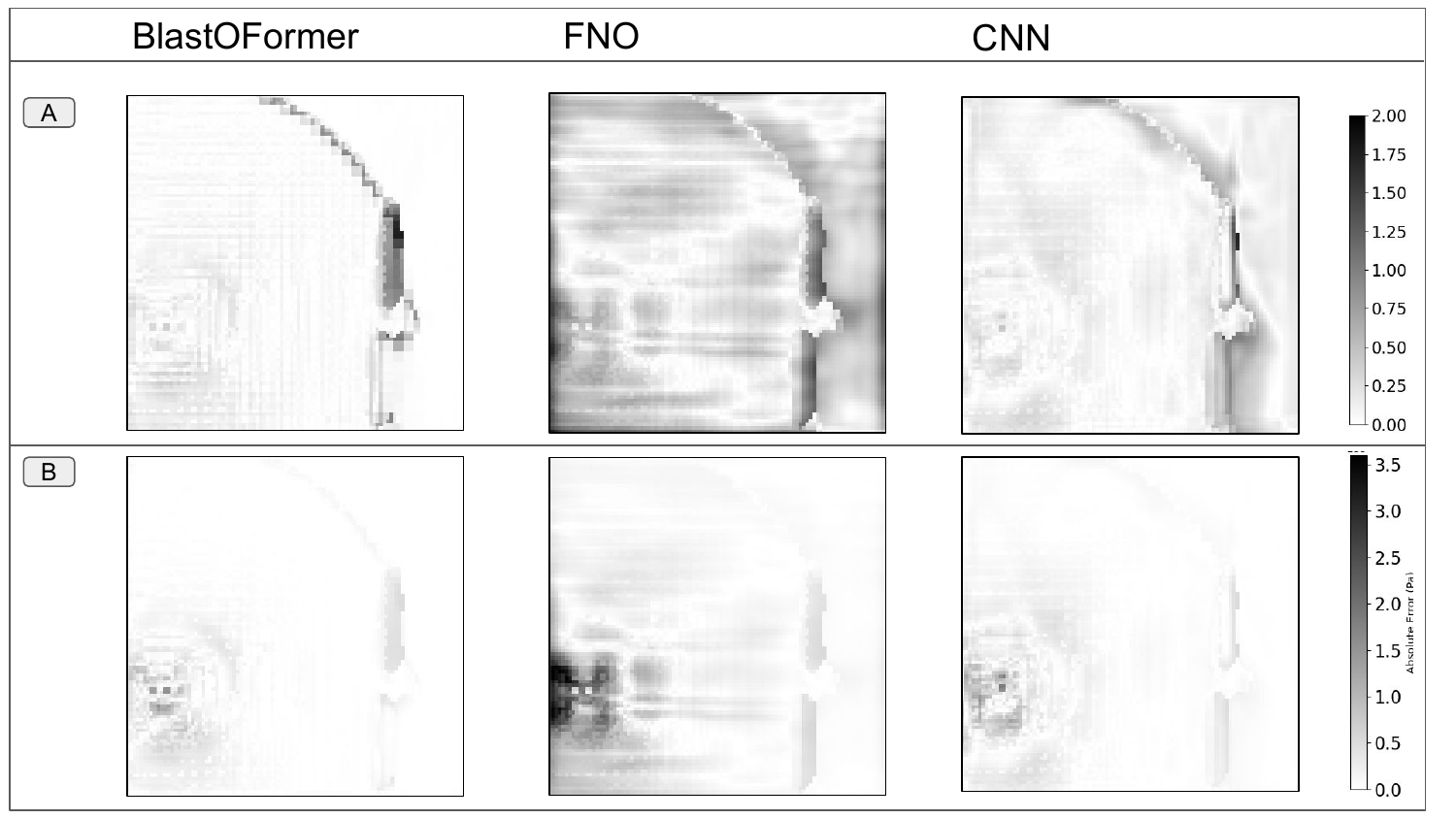}
    \caption{(A) From left to right, absolute error maps comparing predictions from BlastOFormer, FNO, and CNN to the ground truth in the log domain. (B) Corresponding absolute error maps based on unscaled predictions from BlastOFormer, FNO, and CNN. The visualizations utilize a binary colormap, transitioning from white (lower error) to black (higher error). In both visualizations, BlastOFormer demonstrates lower error compared to the FNO and CNN models, indicating superior predictive accuracy.}
    \label{error_maps}
\end{figure}

Figure \ref{error_maps} presents absolute error visualizations comparing predictions from BlastOFormer, FNO, and CNN to the blastFoam ground truth. Subfigure (A) displays error maps in the log transformed domain, while subfigure (B) shows error in the unscaled domain. The binary colormap from white (low error) to black (high error) makes regional differences in predictive accuracy immediately apparent.

In the log-transformed domain, BlastOFormer demonstrates the lowest overall error across the pressure field. The majority of the domain remains near white, indicating minimal deviations from the ground truth. Notably, the largest discrepancies for BlastOFormer appear in regions farther from the charge center, particularly around the central obstacle, where the model underperforms slightly relative to its otherwise consistent output. FNO, on the other hand, displays the highest error throughout most of the domain. While it appears to perform slightly better than BlastOFormer around the middle obstacle, its accuracy degrades significantly elsewhere, especially in peripheral regions. CNN shows a nuanced performance: its error around the middle obstacle is lower than both BlastOFormer and FNO, but it exhibits higher error near the lower obstacle. Visually, CNN appears marginally better than FNO overall in this domain but slightly behind BlastOFormer in terms of global consistency.

In the unscaled domain, differences between the models become even more pronounced. BlastOFormer retains its position as the most accurate model, with error still concentrated around the middle obstacle, but largely reduced across the rest of the domain. This indicates improved magnitude fidelity even in complex regions. FNO again shows the most widespread error, with darker regions clustered near the charge center and elevated error levels surrounding both obstacles. Its failure to match the true pressure magnitudes is more evident here than in the log domain. CNN benefits from the unscaling to some extent, with error levels slightly reduced compared to the log domain. However, it still lags behind BlastOFormer, with prominent dark regions near the charge center and noticeable errors persisting across the field. These observations align with the quantitative results and confirm that BlastOFormer not only preserves spatial structure and magnitude in predictions but also minimizes localized errors more effectively than its counterparts.

\subsection{Error Distribution Analysis}
\begin{figure}[h]
    \centering
    \includegraphics[width=\linewidth]{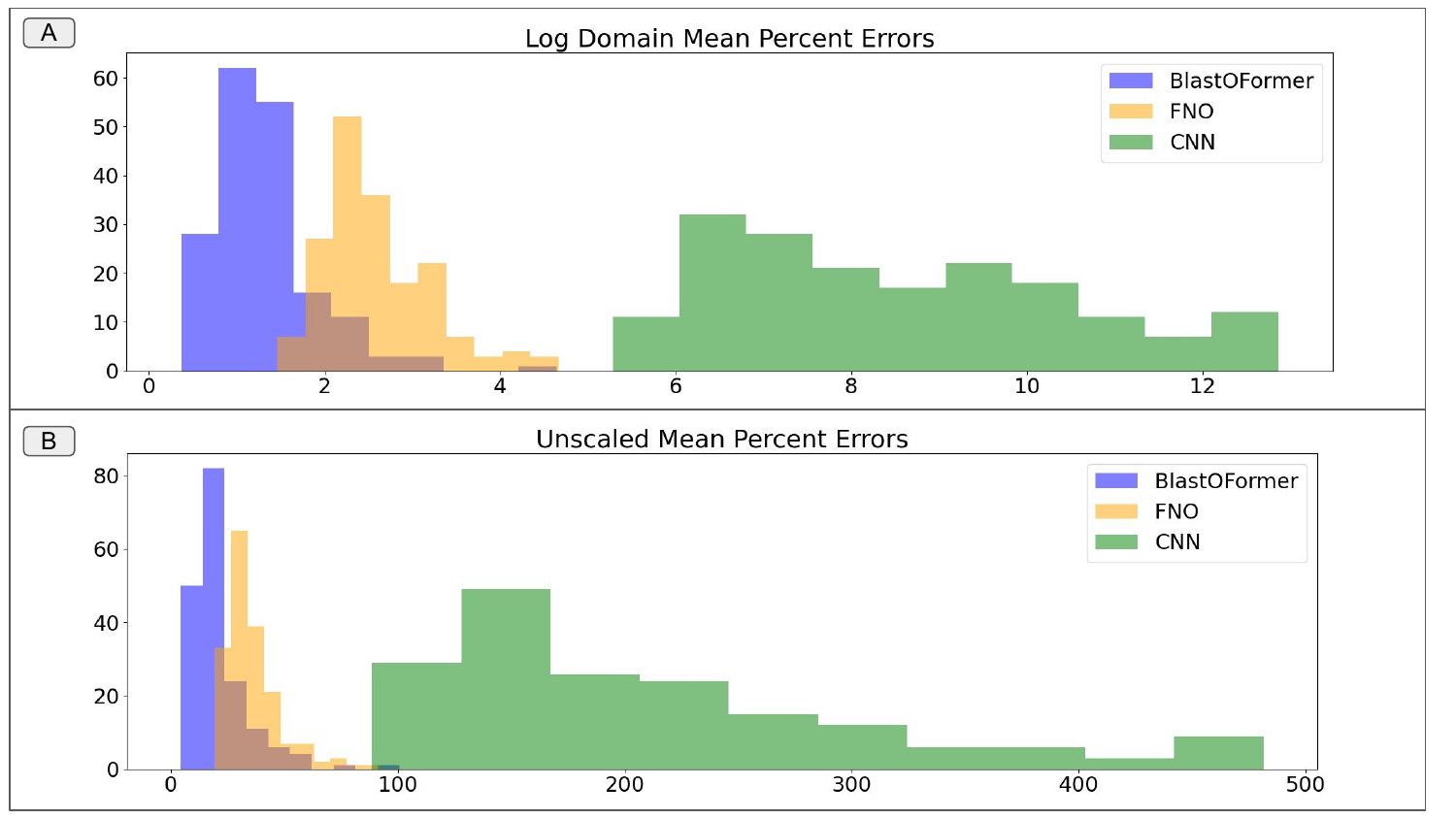}
    \caption{(A) Histogram of the mean percent error for the log domain for BlastOFormer, FNO and CNN. (B) Histograms for. the mean percent error for the unscaled domain.}
    \label{histograms}
\end{figure}
Figure \ref{histograms} presents histograms of the prediction errors for BlastOFormer, FNO, and CNN, comparing model performance in both the log transformed (A) and unscaled (B) domains. These histograms reveal not only central tendencies like mean absolute error but also the spread and concentration of model errors across the entire dataset.

In the log transformed domain (Fig. \ref{histograms}A), BlastOFormer demonstrates the most concentrated error distribution, with the majority of predictions having an MAE near zero and tapering off around 2. A small number of outliers extend slightly above 4\%, but overall the distribution remains tightly clustered in the low error regime. FNO also shows a relatively compact distribution, though it is shifted toward higher error values, generally ranging from just above 1\% to slightly above 4\%. There is partial overlap between FNO and BlastOFormer, but the FNO distribution is broader and less concentrated. In contrast, CNN displays a wide and uniformly distributed error range, spanning from around 5\% up to over 12\%. This distribution indicates a consistently higher prediction error and the absence of a strong mode near low error values, underscoring CNN’s weaker performance in the log domain.

In the unscaled domain (Fig. \ref{histograms}B), a similar pattern is observed, but with magnified error values due to the influence of scale. BlastOFormer again maintains the lowest error distribution, with most predictions exhibiting percent errors below 40. Its distribution remains tightly concentrated, reinforcing its reliability in preserving both spatial and magnitude accuracy. FNO trails behind, with a broader error spread and higher central tendency, although some overlap with BlastOFormer is still present. CNN performs the worst by a wide margin; its error values range dramatically from 100\% to over 500\%, far exceeding the other models. This large spread and skew highlight CNN’s failure to generalize in the unscaled domain, where matching absolute pressure magnitudes becomes critical.

These distributions complement earlier results by illustrating not only which model performs best on average, but also how consistent that performance is across all spatial points. BlastOFormer is not only the most accurate but also the most robust, exhibiting the narrowest and lowest tailed error distributions in both domains.

\section{Conclusion}

In this work, we introduced BlastOFormer, a Transformer based surrogate modeling framework designed to predict maximum blast pressure fields from obstacle and charge configurations. By leveraging a structured grid based representation and integrating advanced attention mechanisms, BlastOFormer effectively captures complex spatial dependencies and nonlinear interactions present in blast wave propagation. The model's architecture, inspired by OFormer and Vision Transformer (ViT), allows it to learn meaningful mappings between environmental conditions and pressure outcomes, while maintaining efficiency and generalizability.

We benchmarked BlastOFormer against established deep learning baselines, including the Fourier Neural Operator (FNO) and a convolutional neural network (CNN), using a dataset generated with high-fidelity blastFoam simulations. Quantitative results showed that BlastOFormer achieved the highest $R^2$, lowest mean absolute error (MAE), and lowest average percentage error across both log transformed and unscaled domains. Qualitative analyses of predicted pressure fields further confirmed its superior ability to capture both the magnitude and spatial structure of pressure distributions. Error maps and histogram analyses demonstrated that BlastOFormer not only produces more accurate predictions, but does so consistently across the entire spatial domain, with lower variance compared to other models.

Importantly, BlastOFormer achieves these results while operating several orders of magnitude faster than traditional CFD solvers, offering a powerful alternative for rapid blast pressure estimation in time critical or resource constrained applications. Its ability to generalize across varying obstacle and charge configurations makes it especially suitable for use in safety assessment, structural design, and mission planning.

Future work may explore extending the model to predict time dependent pressure fields, incorporating 3D spatial data, or adapting it for multi physics scenarios involving thermal effects and material deformation. Additionally, integrating uncertainty quantification could enhance its reliability for real world deployment in risk sensitive domains.



\newpage
\section{Appendix}

\subsection{Additional blastFoam Configuration}

\begin{table}[h]
    \centering
    \begin{tabular}{ccc}
    \hline
      phase   & parameter & value \\
      \hline
       c4 Reactants  & & \\
       & equation of state & Murnaghan \\
       & $\gamma$ & 0.25 \\
       & reference pressure & 101,298 (Pa) \\
       & $\rho_0$ & 1601 ($\frac{kg}{m^3}$) \\
       & $k_0$ & 0 \\
       & molecular weight & 55.0 ($\frac{g}{mol}$) \\
       & $\mu$ & 0 \\
       &  Prandtl number & 1 \\
       & $C_v$ & 1400 ($\frac{J}{kg\cdot K}$)\\
       & $H_f$ & 0.0 ($\frac{kJ}{kg}$)\\
       \hline
       c4 Products & & \\
       & equation of state & Jones-Wilkins-Lee \\
       & $\rho_0$ & 1601  ($\frac{kg}{m^3}$)  \\
       & $A$ & 609.77$\times10^9$ \\
       & $B$ & 12.95$\times 10^9$ \\
       & $R_1$ & 4.5 \\
       & $R_2$ & 1.4 \\
       & $\omega$ & 0.25 \\
       & molecular weight & 55.0 \\
       & $\mu$ & 0 \\
       & Prandtl number & 1 \\
       & $C_v$ & 1400 ($\frac{J}{kg\cdot K}$) \\
       & $H_f$ & 0.0 ($\frac{kJ}{kg}$) \\
       \hline
       c4 & & \\
       & activation model & linear \\
       & $E_0$ & 9.0$\times10^9(m^2\cdot Pa)$  \\
       & $v_{det}$ & 7850 $(\frac{m}{s})$ \\
       & minimum density & $1\times10^{-6} (\frac{kg}{m^3})$ \\
       & minimum volume fraction & $1\times10^{-10}$ (Dimensionless)
    \end{tabular}
    \caption{blastFoam phase properties}
    \label{tab:my_label}
\end{table}
\begin{figure}[h]
    \centering
    \includegraphics[width=\linewidth]{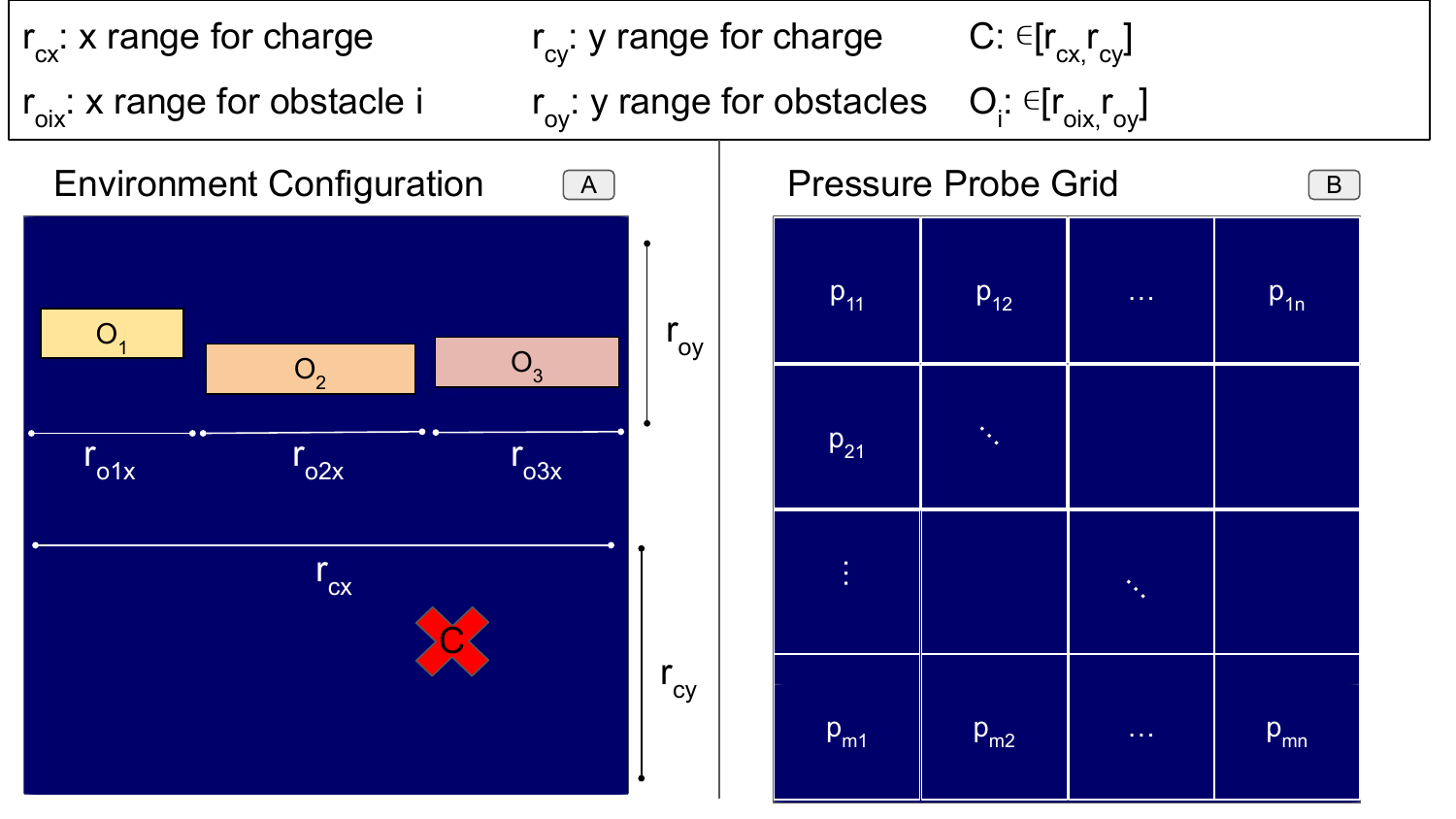}
    \caption{Environment Setup and Pressure Probe Configuration.
    (A) Each dataset sample includes three obstacles randomly positioned within distinct, non overlapping horizontal (x axis) ranges and sharing vertical (y axis) coordinates. The charge mass and location are also randomized within predefined limits, ensuring no spatial overlap between obstacles or between obstacles and the charge mass.
    (B) Pressure probes are uniformly arranged in a structured grid layout throughout the environment, all placed at the same height (constant z coordinate).}
    \label{blastfoam}
\end{figure}

\subsection*{Best Samples}

\begin{figure}[H]
    \centering
    \includegraphics[width=\linewidth]{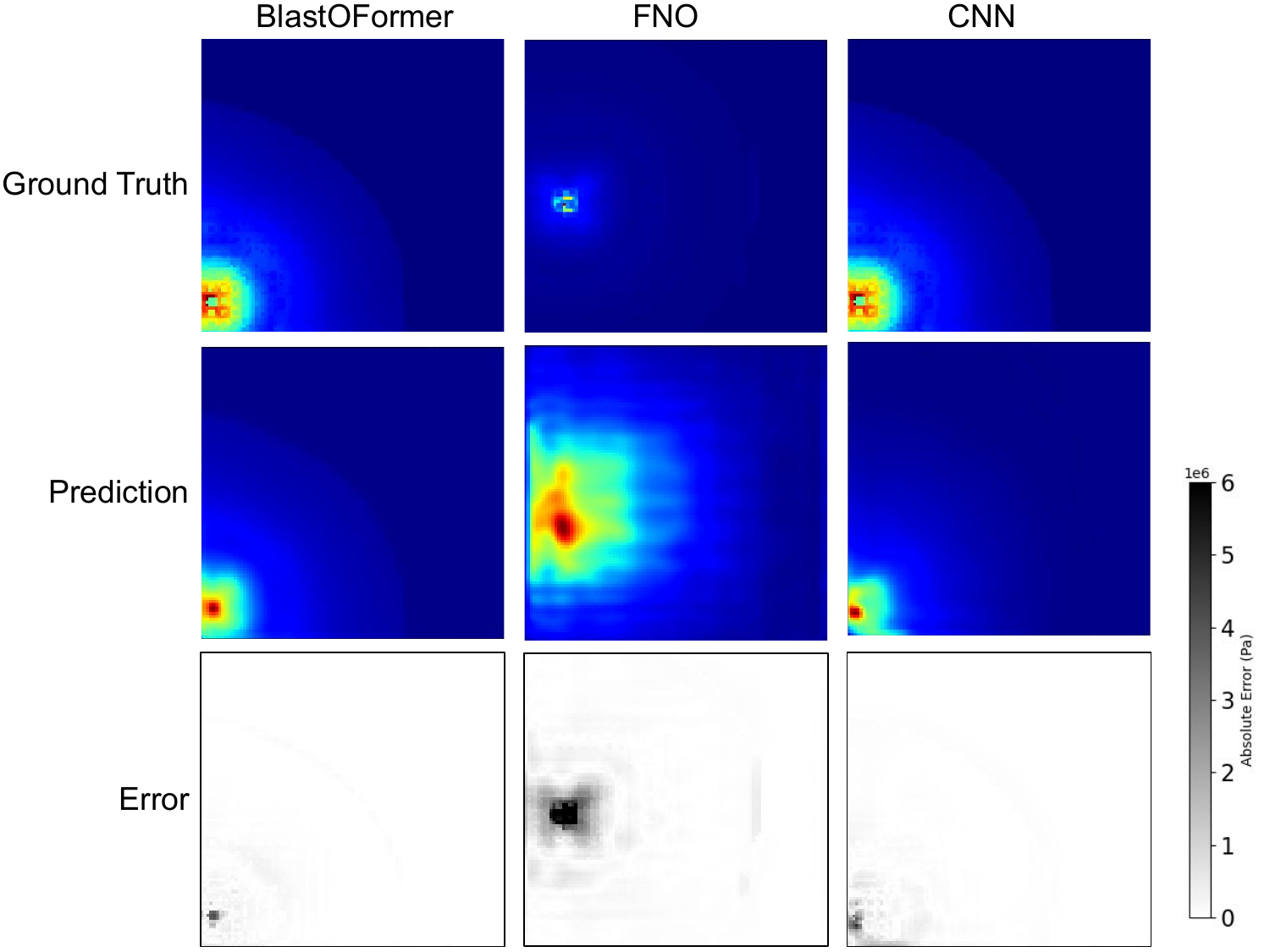}
    \caption{Best performing samples from BlastOFormer, FNO and CNN models}
    \label{Best}
\end{figure}

\subsection*{Worst Samples}
\begin{figure}[H]
    \centering
    \includegraphics[width=\linewidth]{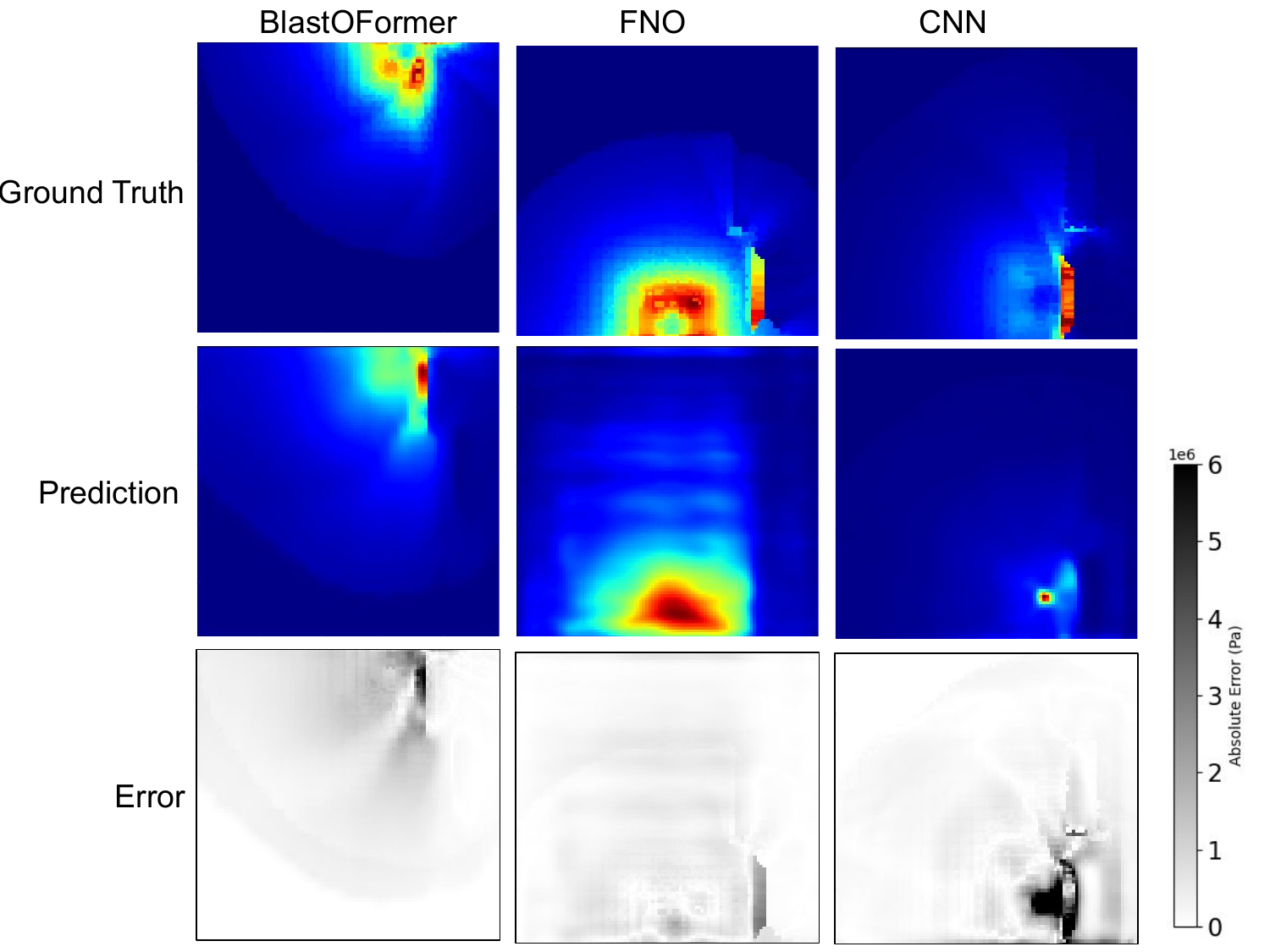}
    \caption{Worst performing samples from BlastOFormer, FNO and CNN models}
    \label{worst}
\end{figure}

\end{document}